\documentclass[10pt,twocolumn,letterpaper]{article}

\usepackage[pagenumbers]{wacv} 

\usepackage{wrapfig}
\usepackage{bm}
\usepackage{enumitem}
\usepackage{graphicx}
\usepackage{amsmath}
\usepackage{amssymb}
\usepackage{booktabs}
\usepackage{subcaption}
\usepackage{tikz}
\usepackage{pifont}
\usepackage{array}
\newcolumntype{P}[1]{>{\centering\arraybackslash}p{#1}}
\usepackage{makecell}
\usepackage{commath}
\usepackage{svg}
\usepackage{graphicx}
\usepackage{balance}
\usepackage{multirow}
\usepackage{tabularx}

\usepackage{algorithm}
\usepackage[noend]{algpseudocode}

\usepackage[pagebackref,breaklinks,colorlinks]{hyperref}

\newcommand{\customsubsection}[1]{%
  \par
  \pagebreak[2]%
  \refstepcounter{subsection}%
    \everypar={%
      {\setbox0=\lastbox}
      \addcontentsline{toc}{subsection}{%
        {\protect\makebox[0.3in][r]{\thesubsubsection.} \hspace*{3pt}#1}}%
      \textbf{\thesubsection\space\space{#1}\space}%
      \everypar={}%
    }%
  \ignorespaces
}

\usepackage[capitalize]{cleveref}
\crefname{section}{Sec.}{Secs.}
\Crefname{section}{Section}{Sections}
\Crefname{table}{Table}{Tables}
\crefname{table}{Tab.}{Tabs.}

\usepackage{xspace}

\newcommand{\ours}{{\fontfamily{cmr}\selectfont Design-o-meter}\xspace}
\newcommand{\ourmethod}{{\fontfamily{cmr}\selectfont SWAN}\xspace}

\graphicspath{ {./images/} }
\makeatletter

\newcommand{\Rmnum}[1]{\expandafter\@slowromancap\romannumeral #1@}
\makeatother

\usepackage[capitalize]{cleveref}
\crefname{section}{Sec.}{Secs.}
\Crefname{section}{Section}{Sections}
\Crefname{table}{Table}{Tables}
\crefname{table}{Tab.}{Tabs.}

\def\wacvPaperID{1545} 
\def\confName{WACV}
\def\confYear{2025}

\begin{document}
\title{\vspace{-9pt}\ours: Towards Evaluating and Refining Graphic Designs}
\author{Sahil Goyal$^{\ast \dagger}$, Abhinav Mahajan$^{\ast \ddagger}$, Swasti Mishra, Prateksha Udhayanan, Tripti Shukla, \\ K J Joseph, Balaji Vasan Srinivasan\\
$^{\dagger}$IIT Roorkee ~~~~~~~ $^{\ddagger}$IIIT Bangalore ~~~~~~~ Adobe Research \\
{\tt\small sahil\_g@ma.iitr.ac.in, abhinav.mahajan@iiitb.ac.in, \{josephkj, balsrini\}@adobe.com}
}
\maketitle
\def\thefootnote{*}\footnotetext{Equal contribution. Work done during internship at Adobe Research.}\def\thefootnote{\arabic{footnote}}

\begin{abstract}
    Graphic designs are an effective medium for visual communication. They range from greeting cards to corporate flyers and beyond. Off-late, machine learning techniques are able to generate such designs, which accelerates the rate of content production. An automated way of evaluating their quality becomes critical. Towards this end, we introduce \ours, a data-driven methodology to quantify the goodness of graphic designs. Further, our approach can suggest modifications to these designs to improve its visual appeal. 
    To the best of our knowledge, \ours is the first approach that scores and refines designs in a unified framework despite the inherent subjectivity and ambiguity of the setting.    
    Our exhaustive quantitative and qualitative analysis of our approach against baselines adapted for the task (including recent Multimodal LLM-based approaches) brings out the efficacy of our methodology. We hope our work will usher more interest in this important and pragmatic problem setting. 
    Project Page: \href{https://sahilg06.github.io/Design-o-meter/}{\tt{sahilg06.github.io/Design-o-meter}}.

  
\end{abstract}

\vspace{-10pt}
\section{Introduction}
\label{sec:intro}

Graphic designs are becoming increasingly ubiquitous: advertisement content,  menu cards at restaurants, campaign flyers, and so on. It is a composite of text, images, and shapes that harmoniously intermingle in an aesthetically pleasing way to convey the intended message effectively.
A typical workflow of a graphic designer involves ideation, creation, and refinement stages. Each of these stages has its unique characteristics: \textit{ideation} involves planning the design, \textit{creation} involves aggregating the design elements and creating the first version, and \textit{refinement} involves improving the design iteratively. The refinement stage is particularly prone to redundancy, as it involves fine-tuning details, adjusting layouts, and sometimes reworking significant portions of the design to meet the desired standards. Designers often undergo numerous feedback and revision cycles, which can be labor-intensive and time-consuming.
Generative AI technologies can work hand-in-hand with designers to supplement them in all phases of their creative workflow.


\begin{figure}
    \centering
 \includegraphics[width=0.48\textwidth]{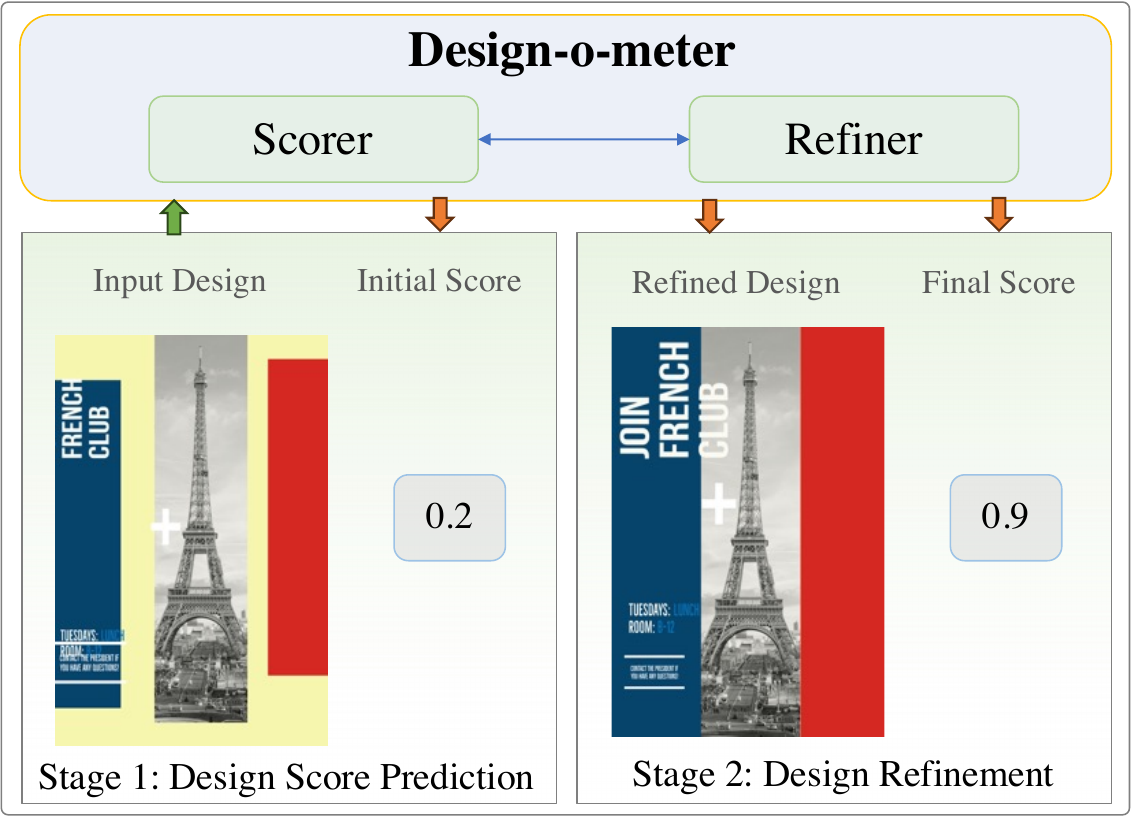}
    \caption{The figure presents an overview of \ours. It takes a design as input. The \textit{scorer} evaluates the design and provides a unified design score. The \textit{refiner} refines the design with the help of the design score to improve its aesthetic appeal.}
    \vspace{-15pt}
    \label{fig:intro}
\end{figure}

Off-late, such generative models \cite{horita2024retrievalaugmentedlayouttransformercontentaware, weng2024desigenpipelinecontrollabledesign, chen2024alignedlayoutgenerationdiffusion, chai2023layoutdmtransformerbaseddiffusionmodel,inoue2023flexiblemultimodaldocumentmodels} has been used for creating designs and layouts from user intents. Some of these methods can take in design assets from a user and generate designs by composing them \cite{horita2024retrievalaugmentedlayouttransformercontentaware, shabani2024visual, hsu2023posterlayout}, while some others can generate the entire graphic design from a text prompt \cite{jia2024colehierarchicalgenerationframework, inoue2024opencolereproducibleautomaticgraphic, yang2023fontdiffuseroneshotfontgeneration}. These can generate large number of designs with very low latency. Coupled with the increasing online and offline consumption of designs, there is a strong demand for automatic tools to evaluate them.

The metrics currently used to evaluate graphic designs, like Fréchet Inception Distance (FID) \cite{heusel2017gans}, mean IOU and max IOU are often insufficient to capture the nuanced aspects of design quality. 
Further, these metrics strongly depend on having a high-quality reference design for comparison. Such evaluations might even penalize the creative freedom of the model (a new design might have a completely different location for the constituent elements but still might look good).
Reference-free evaluation metrics like alignment and overlap \cite{li2020attribute} have also been proposed to evaluate the generated layouts. These metrics fail to provide a holistic assessment of overall design quality. These limitations highlight the need for more comprehensive and robust evaluation methods that can assess the effectiveness, usability, and visual appeal of generated designs such that it corroborates with human perception and its use cases. 



Towards this end, we propose \ours. As shown in \cref{fig:intro}, it contains a \textit{scorer}, which quantitatively evaluates how good a design is, and a \textit{refiner}, which takes in the score from the scorer and refines the design to improve the score. A design is a composite of components like text, images, shapes, and icons. Each component has its properties like color, location, size, content, opacity, shadow, and so on. A visually appealing design has an optimal value for each property and component. Hence, the search space for good designs is indeed combinatorially large.
Further, the design trends also change with time. To effectively model such a complex design space, we propose a data-driven approach for the scorer and a novel meta-heuristic methodology for the refiner. Concretely, the scorer is modeled as a metric learning model that scores good-designs over bad-designs (\cref{sec:design_scorer}), and the refiner is a genetic algorithm with a novel design-specific crossover operation called \ourmethod: De\underline{s}ign Specific Crossover \underline{w}ith Sm\underline{a}rt S\underline{n}apping (\cref{sec:design_refiner}).

We conduct thorough experimental analysis to test the mettle of both our scorer and refiner modules. Our experiments reveal that our scorer is able to capture small nuances in design documents that recent Multimodal LLMs fail to capture, while our newly proposed \ourmethod is able to efficiently navigate the complex design space to refine existing designs, comfortably outperforming recent state-of-the-art approaches. Further, we clearly ablate the different components of the method, and provide sensitivity analysis of the various design choices of our framework. 

\vspace{3pt}
\noindent Our key contributions are summarized below:
\vspace{-3pt}
\begin{itemize}[leftmargin=*]
\setlength\itemsep{-0.05em}
    \item We develop a novel holistic framework, \ours,  that can provide a comprehensive design score and further refine the designs to improve the score.
    \item We propose a metric learning model that learns to disambiguate between good designs and bad designs.
    \item Our novel smart snapping crossover methodology \ourmethod, refines designs to improve its aesthetic appeal.
    \item Our exhaustive quantitative and qualitative evaluation brings out the efficacy of our proposed \ours.
\end{itemize}

\section{Related Work}
In the following subsections, we first discuss related methodologies that evaluate designs and then talk about approaches that refine them. Finally, we provide a brief summary of generic algorithms, which is the basis of our proposed refiner module.

\subsection{Design Evaluation}
Design evaluations strategies can be grouped into heuristic-based approaches and data-driven approaches:

\vspace{-10pt}
\paragraph{Design Heuristics based Approaches:}
\label{para:heuristics}
Design heuristics are cognitive tools that designers and engineers use to measure different design aesthetics. Prior works \cite{ngo2000mathematical, ngo2001another, o2014learning, harrington2004aesthetic, bauerly2006computational, scholgens2016aesthetic, zen2014towards} use heuristics as a fixed set of formulas, each formula measuring a particular aspect of the design. 
Such methods do not consider the overlap between design elements, which is common in graphic designs. Despite the valuable insights provided by the heuristic rules, they don't offer a comprehensive view of the overall design. Also, heuristics are often subjective, and their concrete meaning can deviate, which might lead to inconsistencies.

\vspace{-10pt}
\paragraph{Data-driven Approaches:}

Most of the data-driven approaches \cite{sawada2024visual, dou2019webthetics, wan2021novel} formulate the problem as a regression or classification task, i.e., mapping images to aesthetic ratings given by human annotators. 
Dou \etal \cite{dou2019webthetics} train a convolution network to predict the aesthetic ratings of webpages. 
They use a dataset \cite{reinecke2014quantifying} of webpage screenshots with human-annotated aesthetic ratings. 
Wan \etal \cite{wan2021novel} obtain global, local, and aesthetic features from the webpage layout to predict the aesthetics. 
The above-mentioned techniques rely on human-annotated aesthetic ratings, which have several drawbacks. Creating a high-quality human-annotated dataset is time-consuming and expensive. It may also introduce subjectivity and bias.


An interesting approach that requires minimal or no human interference is the use of Siamese Networks \cite{bromley1993signature}. Ever since they emerged, Siamese Networks have found application in various diverse fields such as self-supervised representation learning \cite{assran2022masked, caron2021emerging, zbontar2021barlow, bardes2021vicreg}, audio-visual synchronization \cite{chung2016out, prajwal2020lip, goyal2023emotionally}, and measuring aesthetics \cite{zhao2018characterizes, kong2022aesthetics++, tabata2019automatic, kong2016photo}, etc. Their popularity is due to several reasons, such as the sharing of parameters among its twin networks, the ability to navigate the search space, and the extraction of distinctive features for downstream tasks. They allow learning in unsupervised settings. They compare the data instances in pairs instead of directly using labels and are robust to data imbalance. 
Zhao \etal \cite{zhao2018characterizes} propose a Siamese-based deep learning framework to estimate personality scores of a graphic design. 
Aesthetics++ \cite{kong2022aesthetics++} train a Siamese-based network to estimate the aesthetics of a design. 
Kong \etal \cite{kong2016photo} and Aesthetics++ \cite{kong2022aesthetics++} use comparatively less human annotation to create training pairs but still introduce bias and may not be accurate. Tabata \etal \cite{ tabata2019automatic} arbitrarily moves the layout elements without any guidance to generate negative examples for training. However, designs generated randomly are not necessarily bad and can sometimes exhibit creative and unique layouts. 
We follow the successful practice of existing works \cite{zhao2018characterizes, kong2022aesthetics++} utilizing Siamese networks. Instead of human annotation and random perturbations without guidance, we use intelligent transformations to create design pairs for training. Moreover, unlike the methods mentioned above, our approach incorporates layout-information-rich color encodings alongside the design renderings, providing a more comprehensive representation of the design elements and additional guidance to the scoring model.

Recently, there are some efforts \cite{jia2023cole,cheng2024graphic} that tries to evaluate a graphic design using large multi-modal LLMs like GPT-4o \cite{OpenAI_2024} and LLaVA \cite{liu2024LLaVAnext}. As there models are trained on huge amounts of data which includes design data too, they are able to analyse graphic designs well. We compare with LLaVA-NeXT and GPT-4o in \cref{sec:exp_score}.

\subsection{Refining Layouts}

Prior works like \cite{o2014learning, o2015designscape} minimize energy functions for typical design principles like white space, symmetry, and alignment using simulated annealing. This approach is not scalable and requires significant computational time (up to 40 minutes to generate one optimized layout), making it impractical for real-time or large-scale applications.
Pang \etal \cite{pang2016directing} generate a set of candidate designs by randomly perturbing the existing design and then selecting the best out of the perturbed designs using heuristic rules. However, such an approach is highly inefficient as the design space is vast and complex. Also, relying on a fixed set of heuristic formulas is not a good approach, as discussed in the section \cref{para:heuristics}. 
Aesthetics++ \cite{kong2022aesthetics++} generates the candidate designs from the input design by traversing a segmentation tree (created using hierarchical segmentation of the graphic design) and leveraging design principles.
The candidate designs with the highest aesthetic score is taken as the refinement suggestion. 
They use a human-annotated dataset for a data-driven approach.
However, the dataset used may not represent all user preferences, leading to biased or skewed results. 
Also, generating multiple candidate designs through tree traversal is computationally intensive and time-consuming.

RUITE \cite{rahman2021ruite} models the refinement as a denoising task and  trains a transformer \cite{vaswani2017attention} model to denoise the layouts.
RUITE only focuses on aligning UI elements; it neglects content and other critical aspects of design, like color schemes and interactive elements. Moreover, reliance on transformer architectures introduces significant latency in the training and inference phases. 
FlexDM \cite{inoue2023flexiblemultimodaldocumentmodels} employs multitask learning in a single transformer-based model to solve various design tasks by predicting masked fields in incomplete vector-graphic documents. While this approach allows for flexible design refinement, it requires significant computational resources, making implementation and scaling challenging. 

In contrast to the above approaches, our method is computationally efficient, providing a more practical solution for design refinement tasks. Our genetic algorithm-based refinement module takes approximately $30$ sec to generate a refined design. We also avoid human intervention in our approach to get unbiased results. 

\subsection{Genetic Algorithms}
Metaheuristic algorithms are applied to address complex real-world problems across domains like engineering, economics, management, etc. 
A genetic algorithm \cite{katoch2021review} (GA) is a metaheuristic algorithm inspired by natural selection.  It is a population-based search algorithm used for optimization. 
Classical GA has an objective function (fitness function), chromosome representation of the population, and operators inspired by biology (selection, crossover, and mutation). Population is improved iteratively using the genetic operators and selecting the fittest.

Multiobjective Genetic Algorithms (MOGAs) differ primarily from standard GAs in how they assign fitness functions, while the remaining steps follow the same procedure as in GAs. They focus mainly on convergence and diversity. NSGA \cite{srinivas1994muiltiobjective}
(Non-dominated sorting genetic algorithm) is a multiobjective genetic algorithm that finds multiple pareto-optimal solutions in a single run. It lacks elitism, needs to specify the sharing parameter ($\sigma_{share}$), and has high computation complexity. To solve these, Deb \etal \cite{996017} introduce a fast, elitist, non-dominated sorting genetic algorithm, NSGA-II. It has been applied in various real-world applications, demonstrating its versatility and effectiveness.

We remove the dependency on multiple objectives as we represent our model score as our unified objective function. We introduce a novel crossover method for designs \ourmethod: De\underline{s}ign Specific Crossover \underline{w}ith Sm\underline{a}rt S\underline{n}apping, and refine the mutation functions for our task. 
With a unified objective function, we bypass the need to filter the pareto-front, which is essential for real user-based scenarios.

\section{Methodology}
Our proposed approach \ours, takes an input a design, and proposes a quantitative score from a \textit{scorer} module, and in-turn uses this score to refine the design using a \textit{refiner} module.
A design is a composite of a set of elements, which are put together in an aesthetic way. Hence, a design $\bm{D}$, can be represented as its metadata $\bm{D}_{meta}$. For instance, a design with a background image and foreground text saying `Hi!', can be represented as $\bm{D}_{meta} = [\{x_1, y_1, w_1, h_1, image\_location\}, \{x_2, y_2, w_2, h_2, `Hi!'\}]$, where $x_i, y_i, w_i, h_i$ refers to the location and dimension information, in its most simplistic form. The image and text can have more attributes like opacity, overlay, shadow, font details, emphasis and so on.
Using $\bm{D}_{meta}$, we can render its rendition image $\mathcal{I}(\bm{D}_{meta})\in \mathbb{R}^{H \times W \times 3}$ and a color encoded layout $\mathcal{L}(\bm{D}_{meta})\in \mathbb{R}^{H \times W \times 3}$, as illustrated in \cref{fig:ce}.
Our scorer $\mathcal{S}(\mathcal{I}(\bm{D}_{meta}), \mathcal{L}(\bm{D}_{meta}))$ learns to measure the goodness and global aesthetics of the input design.
The refinement module $R(\bm{d}_{meta})$, takes as input $\bm{d}_{meta} \subset \bm{D}_{meta}$, which is a set of actionable layout attributes that we are interested in optimizing. It refines these attributes to generate $\bm{{d}_{meta}^{*}}$, by maximizing the corresponding design score from the scorer: $\bm{{d}_{meta}^{*}} \mid \mathcal{S}(I(\bm{{d}_{meta}^{*}}), L(\bm{{d}_{meta}^{*}}) >  \mathcal{S}(I(\bm{d}_{meta}), L(\bm{d}_{meta})) \, \forall \, \bm{d}_{meta}$. $\bm{{d}_{meta}^{*}}$ can be rendered to obtain the refined design. We detail about our scorer and refiner in the next few sub-sections.


\subsection{Design Scorer} \label{sec:design_scorer}
We propose to use a data-driven, self-supervised approach towards learning the scorer module. Different from other works which tries to regress a scalar score from the input design \cite{sawada2024visual, dou2019webthetics, wan2021novel}, we learn a Siamese model in a contrastive setup, to differentiate between good designs and bad designs. 
\vspace{-10pt}
\paragraph{Model Architecture:} 
The scorer function $\mathcal{S}$ is a composition of a feature extractor network $\mathcal{F}$, and a final scoring block $\mathcal{S}_{\textit{block}}$. We use a four layer convolutional network as the feature extractor, and a three layer fully connected network for the scoring block. Given the input design $\bm{D}_{meta}$, its design score is computed as:  
\begin{equation}
    \mathcal{S}(\mathcal{I}(\bm{D}_{meta}), \mathcal{L}(\bm{D}_{meta})) = S_{{block}}(\mathcal{F}(\mathcal{I}(\bm{D}_{meta}), \mathcal{L}(\bm{D}_{meta})))
\end{equation}
\par 
We train our entire model from scratch. Experimentally, we find that using pretrained feature extractors like DINO-V2 \cite{oquab2023dinov2}, CLIP \cite{radford2021learning}, BLIP \cite{li2022blip} and ViT \cite{vaswani2017attention} gives inferior results. This is because these models are trained on natural images, which are distinctly different from the design data, and thus struggle in gauging and extracting important features specific to quantify design aesthetics.

\begin{figure}
    \centering
 \includegraphics[width=0.48\textwidth]{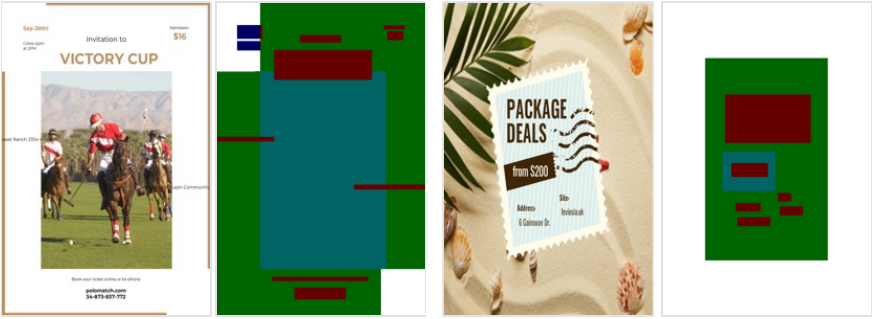}
    \caption{Samples from Crello dataset and their layout encodings.}
    \label{fig:ce}
\end{figure}

Our light-weight convolutional network with group-normalization accepts the rendition image $I(\bm{D}_{meta})$ and the colour-coded layout map $\mathcal{L}(\bm{D}_{meta})$ concatenated across the channel dimension as follows: 
\[ \{N,C,H,W\} + \{N,C,H,W\}  \equiv \{N,2*C,H,W\} \]
where $N$, $C$, $H$, and $W$ are batch size, number of channels, height, and width, respectively.
The layout map $\mathcal{L}(\bm{D}_{meta})$ effectively warps the multi-layer information of a design into the scorer.
\begin{wraptable}[8]{r}{0.3\textwidth}
\vspace{-10pt}
\centering
\caption {Color codes for layout encoding.}
    \label{tab:cs}
\vspace{-10pt}
\resizebox{0.3\textwidth}{!}{%
    \begin{tabular}{lccc}
    \toprule
    Element type & Color (R, G, B) \\
    \midrule
    Image & (0, 100, 0) \\
    Text & (0, 0, 100) \\
    Text and Text overlap & (0, 0, 0) \\
    Text and Image Overlap & (100, 0, 0) \\
    Image and Text Overlap & (100, 100, 0) \\
    Image and Image Overlap & (0, 100, 100) \\
    \bottomrule
    \end{tabular}
\vspace{-25pt}
}
\end{wraptable}
This information helps the model to give attention to the constituent components of the design, and its relative positions while proposing a design score. \cref{tab:cs} summarize the color coding and \cref{fig:ce} shows layout map and its corresponding designs rendition.

\vspace{-10pt}
\paragraph{Training Details:} 
\label{para:train_details}
Similar to the traditional Siamese training setup, the weights of $\mathcal{S}$ is shared. We curate good designs $\bm{D}_{meta}^{{good}}$ and bad designs $\bm{D}_{meta}^{{bad}}$ (explained in the next sub-section) to train the model using the following loss function:
\begin{equation}
\label{eq:combine_loss}
    L_{scorer} = \alpha L_{rank} + \beta L_{sim};
\end{equation} 
where $L_{rank}$ is hinge loss:
\begin{equation}
\label{eq:rankloss}
    L_{rank} = \max(0,~m - (\mathcal{S}(\bm{D}_{meta}^{{good}}) - \mathcal{S}(\bm{D}_{meta}^{{bad}}))
\end{equation}
and $L_{\text{sim}}$ is a similarity loss, formulated as below:
\begin{equation}
    L_{\text{sim}} = \ln(e^{ 2* P_{\text{sim}}(\mathcal{S}(\bm{D}_{meta}^{{good}}),~\mathcal{S}(\bm{D}_{meta}^{{bad}})} + 1); 
\end{equation}
$P_{\text{sim}}$ is an embedding similarity computed as the dot product between the tanh-activations of the ``good" and ``bad" design pairs as follows:
\begin{equation}
    P_{sim} = \frac{ \mathcal{F}(\bm{D}_{meta}^{{good}}).\mathcal{F}(\bm{D}_{meta}^{{bad}})}{\max(\norm{\mathcal{F}(\bm{D}_{meta}^{{good}})}_{2}.\norm{\mathcal{F}(\bm{D}_{meta}^{{bad}})}_{2},\epsilon)} ; \epsilon>0
\end{equation}
\vspace{-10pt}
\paragraph{Dataset Creation:} 
\label{sec:data_create}
Given a set of designs from any design dataset \cite{yamaguchi2021canvasvae}, we first filter them on two criteria: 1) total number of elements should be at most $10$, 2) text should not overlap with images; to create a list of good designs $\bm{D}_{meta}^{good}$.
Next, we surgically modify these designs to make them bad by altering the location and scale of its constituent elements. We employ $22$ such types of transformations to create $\bm{D}_{meta}^{bad}$. The location based perturbations are:

\noindent \textit{Noise addition:} In order to simulate imperfect designs, noise from standard normal distribution with mean $0$ is added to the center coordinates of the design elements. By varying the standard deviation across $0.05, 0.1, 0.2, \text{ and } .5$, we control the degree of perturbation, and thus the degree of badness in the graphic design. 

\noindent \textit{Moving specific type and group of elements differently:} Not all elements of a bad-design will have equal amount of layout shift. In-order to accommodate this aspect, we arbitrarily move specific elements in the graphic design such as the largest element, smallest element, two largest elements, and two smallest elements.

\noindent \textit{Clutter:} Another aspect of designs that makes them naturally bad is clutter, and hence, we clutter all the elements of a graphic design at different positions such as center, top-left, top-right, bottom-left, and bottom-right.

Scale-based transformations are similar to the position-based transformations. The height and width of the design elements are modified instead of the coordinates of the center. Further, the position-based and scale-based transformations are combined to cover more cases. A critical aspect of these augmentations is that they are agnostic to the dataset and are modeled on how humans consider a design good and bad. We can improve these augmentations further to incorporate any newer cognitive constraints, which we leave for future explorations. 
Though the model is trained only on location and scale perturbations, we see in our experiments that the model is able to learn design principles beyond just layout principles. We attribute this to the contrastive learning objective in which the model is trained.

\subsection{Design Refiner}\label{sec:design_refiner}
As a design is a composite of multiple components, search space that constitutes all the designs is vast. Traversing this space of designs in a meaningful way would enable us to find better aesthetically pleasing designs, which improves over the initial design. Here, we propose an efficient approach based on a meta-heuristic algorithm, equipped with our novel \ourmethod: De\underline{s}ign Specific Crossover \underline{w}ith Sm\underline{a}rt S\underline{n}apping. Our scorer (\cref{sec:design_scorer}), guides \ourmethod, to refine its design aesthetics by acting as the objective function being optimized. Our approach expedites convergence and makes the complex algorithm more deterministic.


\begin{figure*}[htbp]
    \centering
    \includegraphics[width=\textwidth]{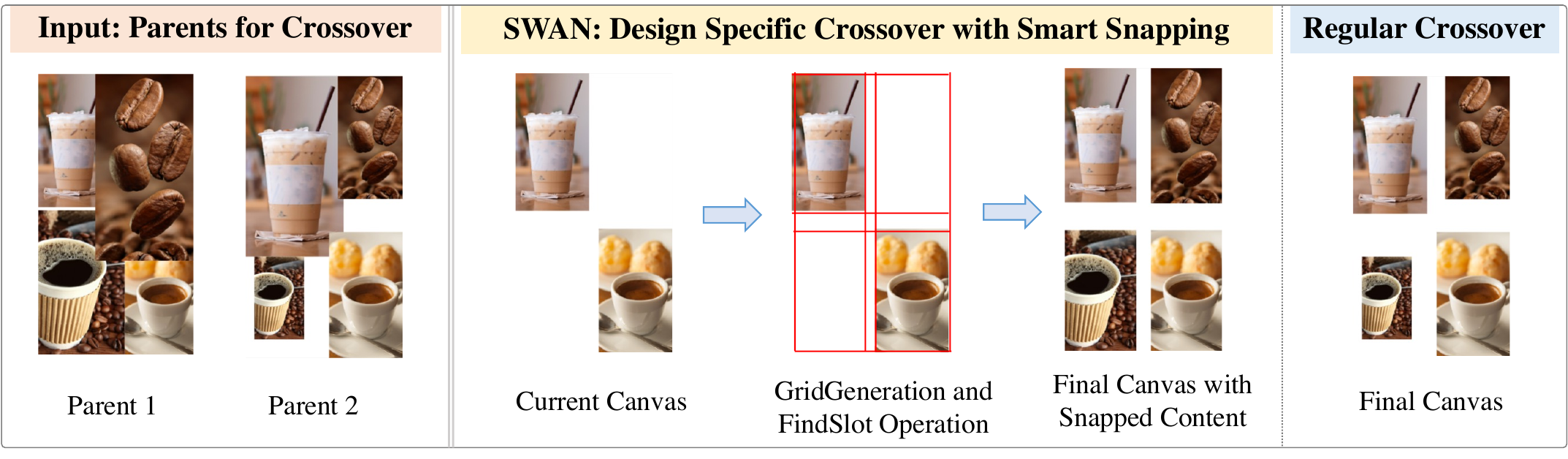} 
    \caption{A visual illustration of \ourmethod: De\underline{s}ign Specific Crossover \underline{w}ith Sm\underline{a}rt S\underline{n}apping.
    Given two parents, \ourmethod first randomly decides which element to pick from either of the parents to generate the child. Then, it copies over the content from the first parent to the current canvas. Next, it identifies potential area within the canvas to host elements from the second parent, guided by grid-lines. Finally, the content from the second parent is `snapped' into the identified areas by changing its attributes. This allows \ourmethod to generate better results when compared with regular crossover. 
    }
    \label{fig:swan}
\end{figure*}
\vspace{-8pt}
\paragraph{Revisiting Genetic Algorithms:} Genetic algorithms are a type of meta-heuristic algorithms used to solve non-linear optimization problems. In our case, we want to optimise the vector $\bm{d}_{meta}^{*}$ on the non-linear objective $\mathcal{S}$, such that the score $\mathcal{S}({\bm{d}_{meta}^{*}})$ is maximised. All genetic algorithms follow a three-step framework: 1) \textit{Initialisation:} where random vectors in the search space ($\bm{d}_{meta}$) are sampled, 2) \textit{Exploitation:} where the ``fitness" of the samples is carried out using the scorer and only the ``fittest" samples are retained, and 3) \textit{Exploration:} where new samples are created with the knowledge of the fittest samples to explore a diverse search space, through crossover and mutation steps.

These steps are carried on until a specified number of iterations or stopping criteria is met. NSGA-II \cite{deb2002fast} introduces a faster filtering operation for selecting the ``fittest" samples.  Deb \etal\cite{deb2013evolutionary} was an augmentation of the previous work and it tailor makes an algorithm specifically for multi-objective optimization. However, many works\cite{ishibuchi2016performance, li2019comparison} argue that this is at the cost of appropriate search space exploration, which is detrimental to overall performance. Since we are explicitly working on single-objective optimization, we use its predecessor. We adapt NSGA-II to our setup by replacing its crossover mechanism with a novel design-specific crossover approach, explained next.

\begin{algorithm}[!t]
\small
\caption{Design Specific Crossover}
\label{algo:training}
\begin{algorithmic}[1]
\Require{Design attributes of Parent 1: $\bm{d}_{meta}^{P1}$; Design attributes of Parent 2: $\bm{d}_{meta}^{P2}$;
}
\Ensure{Design attributes of Child: $\bm{d}_{meta}^{child}$}
\State $\bm{V} \leftarrow \text{Sample}(\bm{d}_{meta}^{P1}, \bm{d}_{meta}^{P2})$ \Comment{\textit{Initialise Parent mapping}}
\State $\bm{L1} \leftarrow \bm{d}_{meta}^{P1}(V == 0)$ \Comment{\textit{Parent1 elements mapping}}
\State $\bm{L2} \leftarrow \bm{d}_{meta}^{P2}(V == 1)$ \Comment{\textit{Parent2 elements mapping}}
\State $\bm{\mathcal{C}} = \bm{L1}$ \Comment{\textit{Initialise Current Canvas}}
\For {$\bm{Element} \in L2$}
\If{$Type(\bm{Element})\text{ == Image or SVG}$}
\State $\bm{G} \leftarrow GridGeneration(\bm{\mathcal{C}})$ \Comment{\textit{Grid Line Generation}}
\State $\bm{Slot} \leftarrow FindSlot(\bm{G})$ \Comment \textit{Find Appropriate Slot}
\State $\bm{{Elem}^{*}} \leftarrow Snap(\bm{Element}, \bm{Slot})$ \Comment{\textit{Refinement}}
\ElsIf{$Type(\bm{Element})\text{ == Text}$}
\State $\bm{E} \leftarrow EulerDistances(\bm{Element}, \bm{\mathcal{C}})$
\State $\bm{{Ele}_{\textit{min}}} \leftarrow MapElement(\bm{\mathcal{C}}, min(\bm{E}))$
\State $\bm{{Elem}^{*}} \leftarrow MinAlign(\bm{Element}, \bm{{Ele}_{\textit{min}}})$
\EndIf
\State $\bm{\mathcal{C}} \leftarrow \bm{\mathcal{C}}{\textit{.append}}(\bm{{Elem}^{*}})$ \Comment{\textit{Update Canvas}}
\EndFor
\State $\bm{d}_{meta}^{child} = \bm{\mathcal{C}}$
\end{algorithmic}
\end{algorithm}

\vspace{-8pt}
\paragraph{\ourmethod: De\underline{s}ign Specific Crossover \underline{w}ith Sm\underline{a}rt S\underline{n}apping} An overview of the approach is presented in \cref{algo:training} and \cref{fig:swan}. In line $1$, two parents (vectors in the search space, $\bm{d}_{meta}^{P1}$, $\bm{d}_{meta}^{P2} \in \bm{d}_{meta})$ are randomly chosen from the set of ``fit" samples (samples remaining after the exploitation step). The objective of crossover function is to fuse the knowledge of both the parents to create a ``smarter" offspring, which becomes a new sample in the population. In regular crossover, there is blind copy-pasting of parent data, and we find that to be unfit for our task at hand, which motivates us to propose \ourmethod: De\underline{s}ign Specific Crossover \underline{w}ith Sm\underline{a}rt S\underline{n}apping. The first step is to initialize a random mask, $\bm{V}$, which assigns what parent is responsible for the percolation of which element in the new Child vector ($\bm{d}_{meta}^{child}$). Next, we copy element attributes from the first parent $\bm{L1}$ using the mask $\bm{V}$ and build our current canvas $\bm{C}$ in Line 4 and the third column in \cref{fig:swan}. For the remaining image elements from the second parent, we build grid lines from the edges of the design elements inside the current canvas and find all the boxes thus made after the intersection of the lines. We then find the box with the most similar size and proximity to the element from Parent2, and then `Snap' it inside it and update the Canvas as shown in \cref{fig:swan}.
For text elements, we find the Euler distances from the center of the text and all the other centers of elements from the current canvas. We find the element with least distance and find out what the cost (distance moved) is of either x-aligning or y-aligning with the element. We choose the minimum of it and update the current canvas to continue.

The mutation step in NSGA-II can also be improved. Standard mutation operation, which adds Gaussian noise to elements, might result in partial overflow in the design elements of their canvas size. We identify such cases and add noise only if it is within the canvas area. This allows our method to skip over degenerate cases. 

\section{Experiments and Results}

\paragraph{Datasets:} \label{para: exp_data} We utilize the widely-used Crello \cite{yamaguchi2021canvasvae} dataset for our evaluation. It contains $23182$ design templates. The data is divided into $18768$ / $2315$ / $2278$ examples for train, validation, and test splits. 
We use multiple guided perturbations techniques to make the design pairs as discussed in \cref{sec:data_create}. 
If the evaluation dataset includes the same or similar transformations used in training, we call it a \emph{biased} setting; otherwise, it is called \emph{unbiased}. We employ two \emph{unbiased} setups: color and cross-match. In the color dataset, the transformation used makes the design's color scheme bad by recoloring certain elements of it. 
We traverse the design, find overlapping elements (say, SVG  preceding a Text), we extract the colors: $\mathcal{C}_\textit{SVG}$, $\mathcal{C}_\textit{text}$ and randomly choose one of the elements(say, the SVG) and recolor it such that the \textit{CIELAB\_Distance}($\mathcal{C}_{\textit{text}}$, $\mathcal{C}_{\textit{SVG}}^{*}$) $\in$ (2, 3). This ensures that there are distinct colours but it makes the design unaesthetic. 
In the cross-match dataset, we randomly pair a \emph{good} version of a design x with the \emph{bad} version of a design y, where x$\neq$y.

\vspace{-5pt}
\paragraph{Baselines:} We compare our \textit{scorer} with recent Multimodal LLM-based design scorers. Specifically, we compare against GPT-4o \cite{OpenAI_2024} and LLaVA-NeXT \cite{liu2024LLaVAnext}. 
We compare the \textit{refiner} against leading design refinement approaches. Specifically, we compare against SmartText++ \cite{li2021harmonious}, FlexDM \cite{inoue2023flexiblemultimodaldocumentmodels}, and COLE \cite{jia2024colehierarchicalgenerationframework} for text box placement (refine-text setting); and against CanvasVAE \cite{yamaguchi2021canvasvae}, FLexDM \cite{inoue2023flexiblemultimodaldocumentmodels}, DocLap \cite{zhu2024automatic}, GPT-4 \cite{openai2024gpt4technicalreport}, and GPT-4V \cite{OpenAI_2023} for full design refinement (refine-all setting).

\vspace{-5pt}
\paragraph{Implementation Details:}
We employ a learning rate of $1e^{-4}$, the Adam optimizer \cite{duchi2011adaptive} with gradient coefficients of [$0.5$, $0.99$], L2 regularization with weight decay as $0.005$ and every $5$ epochs, we schedule the learning rate to divide by half. We set a hard margin, $m$ = $0.2$, and loss parameters $\alpha$ = $0.8$ and $\beta$ = $0.2$ (see \cref{eq:combine_loss}). We add sensitivity analysis on these hyper-parameters in \cref{sec:sensitivity}. 
All the convolution layers in the scorer model are as follows (representing the number of filters, kernel size, and stride, respectively): $(64, 3, 1)$.
We use group-normalization \cite{wu2018groupnormalization} (ngroup=2) in our scoring module. 
For \ourmethod, we find that $\textit{population\_size}=100$, $\textit{n\_trials} = 1500$, $\textit{p} =0.3$ gives good performance. 

\vspace{-5pt}
\paragraph{Evaluation Metrics.}
We introduce Rank Accuracy (RAcc) to evaluate the design scoring methods. The higher the RAcc, the better the scorer. We define the Rank Accuracy as follows:

\begin{equation}
    \text{RAcc} = \frac{1}{N} \sum_{i=1}^{N} \mathbb{I}\{ \mathcal{S}(g_i) > \mathcal{S}(b_i) \};
\end{equation}
If the score given by a scorer $\mathcal{S}$ to a \emph{good} design ($g_i$) is more than that of a \emph{bad} design ($b_i$) in a ($g_i$,$b_i$) pair, we consider it as correctly classified. 

Following FlexDM \cite{inoue2023flexiblemultimodaldocumentmodels}, we use mean Intersection over Union (mIOU) and mean boundary displacement error \cite{li2021harmonious} (mBDE) to evaluate our refinement module for refine-text experiment. Following CanvasVAE \cite{yamaguchi2021canvasvae}, we use Type-wise mIoU (T-mIoU) to evaluate our refinement module for refine-all settings. In T-mIoU, first we calculate the overall IoU for each type of design element (text, image, SVG). Then we take the average of IoU of all the types.
 \begin{equation}
 \vspace{-5pt}
     \text{T-mIoU} = \frac{1}{N} \sum_{i=1}^{N} \frac{\sum_{j=1}^{n_{i}} x_{ij}\cap y_{ij} }{\sum_{j=1}^{n_{i}} x_{ij}\cup y_{ij} };
 \end{equation}
where element $x_{ij}$ belongs to type $i$ and is $j$th element of the ground truth design.  $y_{ij}$ belongs to type $i$ and is the $j$th element of refined design. $n_{i}$ is total number of elements of type $i$ and $N$ is total number of types.

\subsection{Evaluating the Scorer}
\label{sec:exp_score}
We use three settings (biased, unbiased color, and unbiased cross-match) for evaluation. Each data instance in these datasets is a pair of designs, where the task of the model is to provide a better score to the \emph{good} design. 

\begin{wraptable}[12]{r}{0.25\textwidth}
\vspace{-10pt}
\centering
\caption{
Comparison with LMM evaluators in biased, unbiased color (UC) and unbiased cross-match (UCM) setting.}
\label{table:accuracy}
\vspace{-10pt}
\resizebox{0.25\textwidth}{!}{%
   \begin{tabular}{clcc}
\toprule
Setting & Model & RAcc ($\uparrow$) & Params\\ 
\midrule
\multirow{3}{*}{Biased} & LLaVA-NeXT & 34.17 &  7B\\ 
& GPT-4o & \underline{68.84} & -\\
 & \ours & \textbf{94.97} &  $\sim$410k  \\ 
\midrule
\multirow{3}{*}{UC} & LLaVA-NeXT & 44.22 & 7B \\ 
& GPT-4o & \underline{72.86}&  -\\
 & \ours & \textbf{90.45}& $\sim$410k \\
\midrule
\multirow{3}{*}{UCM}  & LLaVA-NeXT & 31.00& 7B \\
& GPT-4o & \underline{63.50}&  -\\ 
 & \ours & \textbf{87.50}& $\sim$410k \\
\bottomrule
\end{tabular}
\vspace{-30pt}
}
\end{wraptable}
We use the following prompt for the LLM-based evaluators: "\textit{I will show you two designs, and you should give each design a design score between 1-100, which follows design principles and reason, and justify your score briefly and succinctly and then output which design has the higher score}". See \cref{table:accuracy} for results. Despite having only $\sim$410k parameters compared to the billions in multi-modal LLMs, our model achieves the highest RAcc across all three datasets. 
This shows that multi-modal LLM-based evaluation of graphic design \cite{jia2023cole,cheng2024graphic} is not good at gauging micro-aesthetic differences between designs \cite{haraguchi2024gptsevaluategraphicdesign}. 

\begin{table*}[h]
\caption{Comparison with state-of-the-art approaches when all elements of a design are refined (Refine-All setting).}
\label{tab:full_gen1}
\centering
\resizebox{0.99\textwidth}{!}{%
\begin{tabular}{lcccccccc}
\toprule
& CanvasVAE & FlexDM & DocLap & GPT-4 0-shot & GPT-4 1-shot & GPT-4V 0-shot & GPT-4V 1-shot & \ours \\
\midrule
T-mIOU ($\uparrow$) & 42.39 & \underline{50.08} & 43.75 & 30.75 & 29.97 & 28.81 & 35.17 & \textbf{54.44} \\
\bottomrule
\end{tabular}
}
\end{table*}
\begin{figure*}
    \centering
 \includegraphics[width=0.99\textwidth]{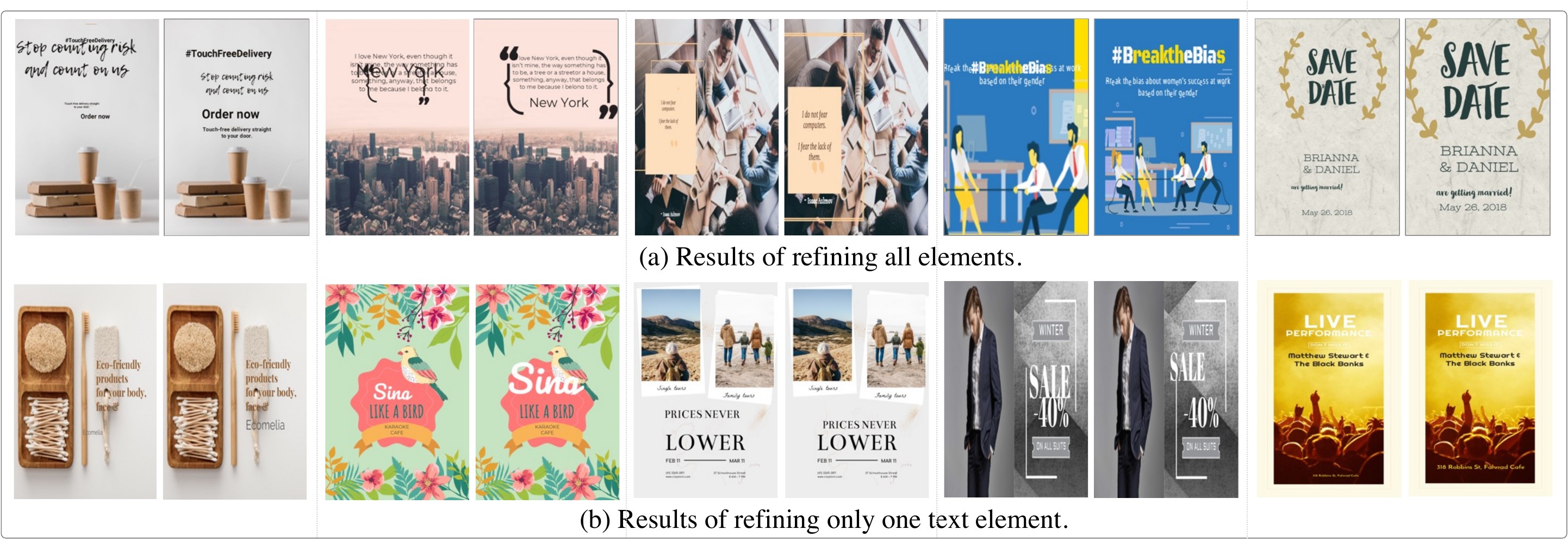}
    \caption{
    Qualitative results of \ours refining graphic designs. The top sub-figure refines all elements of a design (Refine-All setting), while the bottom one refines a single text box (Refine-Text setting). We see that our approach is able to automatically improve the position and scale parameters of design elements, making them more visually appealing.
    }
    \label{fig:final_main_results}
\end{figure*}


\subsection{Evaluating the Refiner}
Following earlier works \cite{inoue2023flexiblemultimodaldocumentmodels, jia2024colehierarchicalgenerationframework}, we consider two settings: refining a single text box and refining whole design.

\begin{table}[H]
\caption{When compared to state-of-the-art approached that refines placement of text on a design, \ours outperforms them in SingleText and MultipleText settings. Further, we ablate \ourmethod to find that it significantly helps to improve performance.
}
\label{tab:tbp}
\centering
\resizebox{0.48\textwidth}{!}{%
\begin{tabular}{lcc cc}
\toprule
& \multicolumn{2}{c}{SingleText} & \multicolumn{2}{c}{MultipleText} \\
\cmidrule(lr){2-5}
& mIoU ($\uparrow$) & mBDE ($\downarrow$) & mIoU ($\uparrow$) & mBDE ($\downarrow$) \\
\midrule
SmartText++ & 0.047 & 0.262 & 0.023 & 0.300 \\
FlexDM & 0.357 & \underline{0.098} & 0.110 & 0.141 \\
COLE & \underline{0.402} & - & 0.172 & - \\
\midrule
Ours w/o \ourmethod & 0.376 $\pm$ 0.01 & 0.116 $\pm$ 0.01 & \underline{0.311} $\pm$ 0.05 & \underline{0.128} $\pm$ 0.01\\
\ours & \textbf{0.42} $\pm$ 0.01 & \textbf{0.08} $\pm$ 0.01 &\textbf{ 0.38} $\pm$ 0.05 & \textbf{0.06} $\pm$ 0.01\\
\bottomrule
\end{tabular}
}
\end{table}
\paragraph{Text Box Refinement (Refine-Text):}
We divide the experiment into two settings: SingleText and MultiText.
SingleText refers to the evaluation of designs containing only one text element, whereas MultiText refers to the evaluation of designs containing multiple text elements. The task is to predict the correct position and size of the target text element, keeping the aspect ratio of the element fixed. Our settings for the experiment are as follows: We choose a random text element as the target element. We randomly initialize the values of the center coordinates and scale of the target. We optimize these values using our refinement module.
We summarize the quantitative results in table \cref{tab:tbp}. 
In the second-last row, we selectively turn-off \ourmethod from \ours. The experiment reveals that our approach outperforms state-of-the-art approaches, and showcases the efficacy of \ourmethod.
We add qualitative results in \cref{fig:final_main_results} (b).


\paragraph{Full Refinement (Refine-All):} In this experiment we randomly initialize the values of the center coordinates and scale of all elements except the background element. 
\cref{tab:full_gen1} showcases comparison with $7$ baseline approaches, and \ours comfortably outperforms all of them. We add qualitative results in \cref{fig:final_main_results} (b).


\begin{figure}
    \centering
 \includegraphics[width=0.5\textwidth]{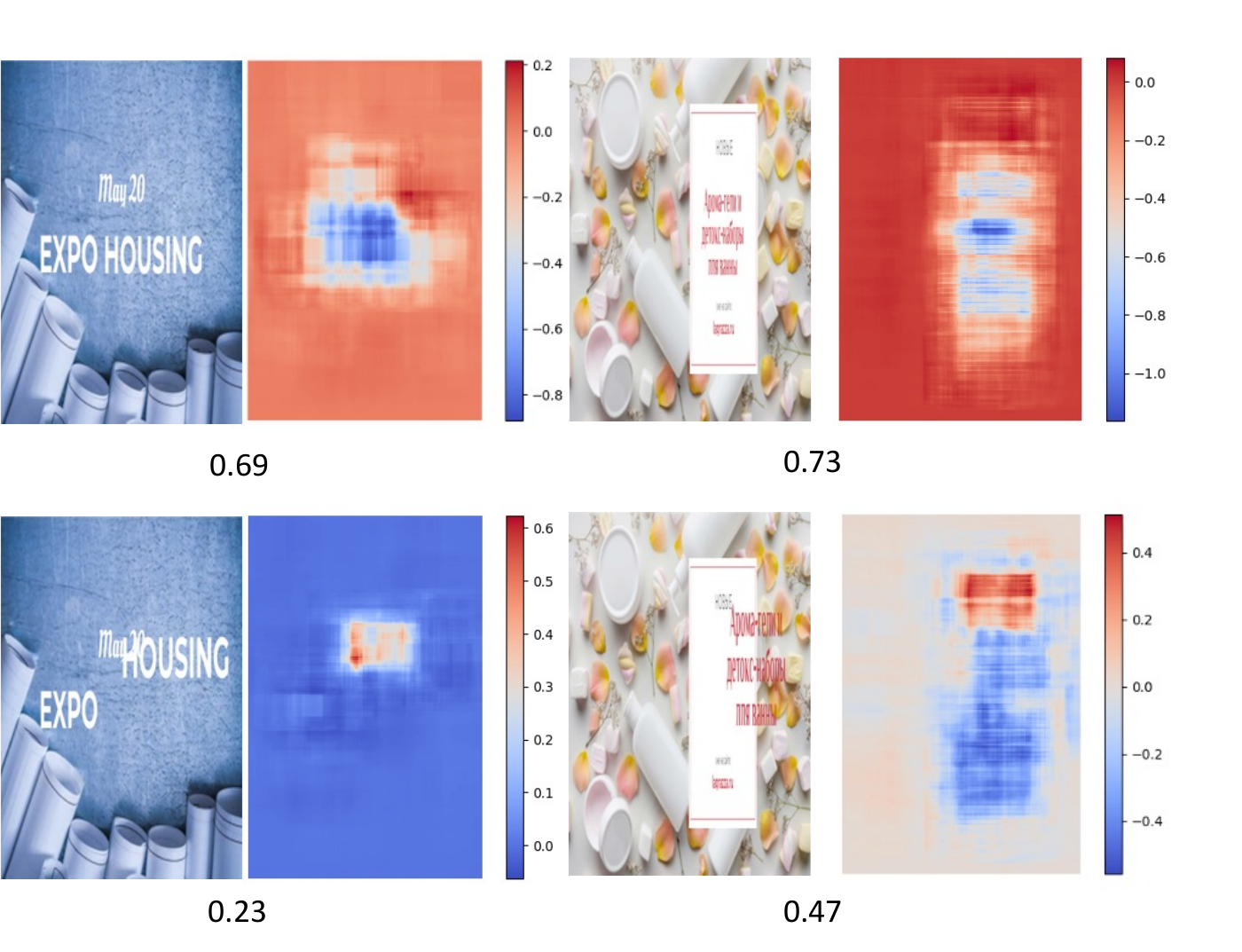}
 
    \caption{
    These occlusion-based sensitivity maps show the locations of the design with positive (red) or negative (blue) impacts on a design score prediction. The numbers correspond to scores.}
    \label{fig:sens}
\end{figure}



\section{Discussion and Analysis}

\customsubsection{Visualizing the \textit{scorer}:} We conduct a simple occlusion based sensitivity analysis to measure the contribution of each region of a design towards the design score. 
We slide a $60 \text{ px} \times 60 \text{ px}$ occluding window across the design, replacing the pixels within the window with the design's mean pixel value at each location. This modified design is then evaluated to generate a new score. The difference between the new score and the original score is used as the pixel value for the window's center in the sensitivity map. This process creates a pixel-wise sensitivity map highlighting areas of the design that most positively or negatively impact the design score. The visualization in \cref{fig:sens} shows that the scorer is indeed paying attention to the key areas of the design while making its decision.
\customsubsection{On the Efficacy of Layout Encoding:}
Including layout color encoding in the input to the scorer model makes it easy for the model to capture the basic design principles such as overlap and alignment.
We experiment with three configurations to verify this hypothesis: only rendering, only layout, rendering $+$ layout; and train the model on Crello dataset. We notice that using rendering $+$ layout as input performs the best with RAcc of {94.97} while the other two configurations give $86.65$ and $86.99$ RAcc.

\customsubsection{Effect of Normalization Layers:}
Every graphic design is unique and complex. Assuming that designs belong to a common distribution and using normalization methods like BatchNorm \cite{ioffe2015batchnormalizationacceleratingdeep} degrades the performance of the model.
\begin{wraptable}[7]{r}{0.25\textwidth}
\vspace{-10pt}
\centering
\caption{We find that GroupNorm gives the best results with the scorer.}
        \label{tab:norm}
\vspace{-10pt}
\resizebox{0.25\textwidth}{!}{%
        \begin{tabular}{lc}
            \toprule
            Method & RAcc ($\uparrow$)\\
            \midrule
            CNN + BatchNorm & 68.10 \\
            CNN + LayerNorm & 82.36 \\
            CNN + InstanceNorm & 85.18 \\
            CNN + GroupNorm & \textbf{94.97} \\
     
       \bottomrule
        \end{tabular}
\vspace{-25pt}
}
\end{wraptable}
We experiment with BatchNorm \cite{ioffe2015batchnormalizationacceleratingdeep}, InstanceNorm \cite{ulyanov2017instancenormalizationmissingingredient}, LayerNorm \cite{ba2016layernormalization}, and GroupNorm \cite{wu2018groupnormalization}.
Table \cref{tab:norm} shows the results. We observe that GroupNorm performs best along with the rendering$+$layout as input.

\customsubsection{Sensitivity Analysis:} \label{sec:sensitivity}
We experiment with different types of margins in \cref{eq:rankloss}. 
\begin{wraptable}[6]{r}{0.13\textwidth}
\vspace{-13pt}
\centering
\caption{Varying $\alpha$ and $\beta$.}
        \label{tab:methods}
\vspace{-10pt}
\resizebox{0.13\textwidth}{!}{%
        \begin{tabular}{@{}lcc@{}}
            \toprule
            ${\alpha}$ & {$\beta$} & RAcc$\uparrow$ \\
            \midrule
            1.0 & 0.0 & 91.50 \\
            0.9 & 0.1 & 93.90 \\
            \textbf{0.8} & \textbf{0.2} & \textbf{94.97}
            \\
            \bottomrule
        \end{tabular}
\vspace{-25pt}
}
\end{wraptable}
Hard margin (H-Margin), transformation-based margin (TB-Margin), and adaptive margin (Ada-Margin) gave $94.97$, $94.10$ and $93.65$ RAcc values. We use H-Margin for the results. Next, we analyze the contribution of $\alpha$ and $\beta$ parameters in \cref{eq:combine_loss}. Decreasing $\alpha$ and increasing $\beta$, seems to have a positive effect, and hence we use $\alpha=0.8$ and $\beta=0.2$ for all experiments.

\section{Conclusion}
In this paper, we propose \ours, a novel framework which can evaluate and refine graphic designs. It takes an existing user design as input, and provide a unified design score. Further, it refines the input design to improve the design score. Our exhaustive experimental analysis brings out the efficacy of \ours.
It would be interesting to see how our scorer can aid other layout and design generation frameworks as an off-the-shelf discriminator. We leave this for future exploration. 
We hope \ours will kindle interest in this practical and relevant setting of evaluating and refining graphic designs. 


{\small
\bibliographystyle{ieee_fullname}
\bibliography{egbib}
}
\newpage
\appendix

\section{Additional Ablation Experiments}
\subsection{Alternate Margins for Eqn.~3} We experiment with Hard margin (H-Margin), transformation-based margin (TB-Margin), and adaptive margin (Ada-Margin). For H-Margin, we try low ($0.2$), medium ($0.5$), and high ($1.0$) values. A low value of $0.2$ achieves the highest RAcc of $\textbf{94.97}$. For TB-Margin, we assign different margins to different transformations, a low value ($0.2$) for transformations adding noise with a low std deviation, and similarly for medium ($0.4$) and high ($0.6$) noises. We define Ada-Margin as the maximum Euclidean distance between feature embeddings of designs in a pair across a batch, represented as:

\begin{equation*}
    \text{Ada-Margin} = \max\left( \max_{(g_i, b_i) \in \text{batch}} \lambda\norm{\mathcal{F}(g_i)-\mathcal{F}(b_i)}_{2}, 0.1 \right)
\end{equation*}

where $\lambda$ is a scaling factor, we choose $\lambda=0.05$.

\subsection{Alternate Similarity Loss Formulation}
We experiment with different similarity losses (exponential, binomial deviance, and square) for Eqn.~4.

\begin{align}
    L_{\text{sim}}^{\text{exp}}&=e^{P_{\text{sim}}(\mathcal{S}(\bm{D}_{meta}^{{good}}),~\mathcal{S}(\bm{D}_{meta}^{{bad}}))} \\
    L_{\text{sim}}^{\text{dev}} &= \ln(e^{ 2* P_{\text{sim}}(\mathcal{S}(\bm{D}_{meta}^{{good}}),~\mathcal{S}(\bm{D}_{meta}^{{bad}})} + 1) \\
    L_{\text{sim}}^{\text{sq}} &=(P_{\text{sim}}(\mathcal{S}(\bm{D}_{meta}^{{good}}),~\mathcal{S}(\bm{D}_{meta}^{{bad}})) + 1)^2
\end{align}

$P_{\text{sim}}$ is an embedding similarity computed as the dot product between the tanh-activations of the ``good" and ``bad" design pairs as follows:
\begin{equation}
    P_{sim} = \frac{ \mathcal{F}(\bm{D}_{meta}^{{good}}).\mathcal{F}(\bm{D}_{meta}^{{bad}})}{\max(\norm{\mathcal{F}(\bm{D}_{meta}^{{good}})}_{2}.\norm{\mathcal{F}(\bm{D}_{meta}^{{bad}})}_{2},\epsilon)} ; \epsilon>0
\end{equation}

We get similar RAcc using all the similarity losses with minor differences. We choose $L_{\text{sim}}^{\text{dev}}$ because of the validation loss decreases most in this case.

\subsection{Use of Classifier Guidance in Scorer} 
We try to guide the scorer model with a binary classification head on top of the siamese model. We additionally introduce a binary cross-entropy loss to differentiate good and bad designs. This setting increases the model size and training time, but doesn't significantly help the scorer model in ranking the designs better.

\section{Failure Cases}
\begin{figure}[H]
\vspace{-10pt}
    \centering
\includegraphics[width=0.5\textwidth]{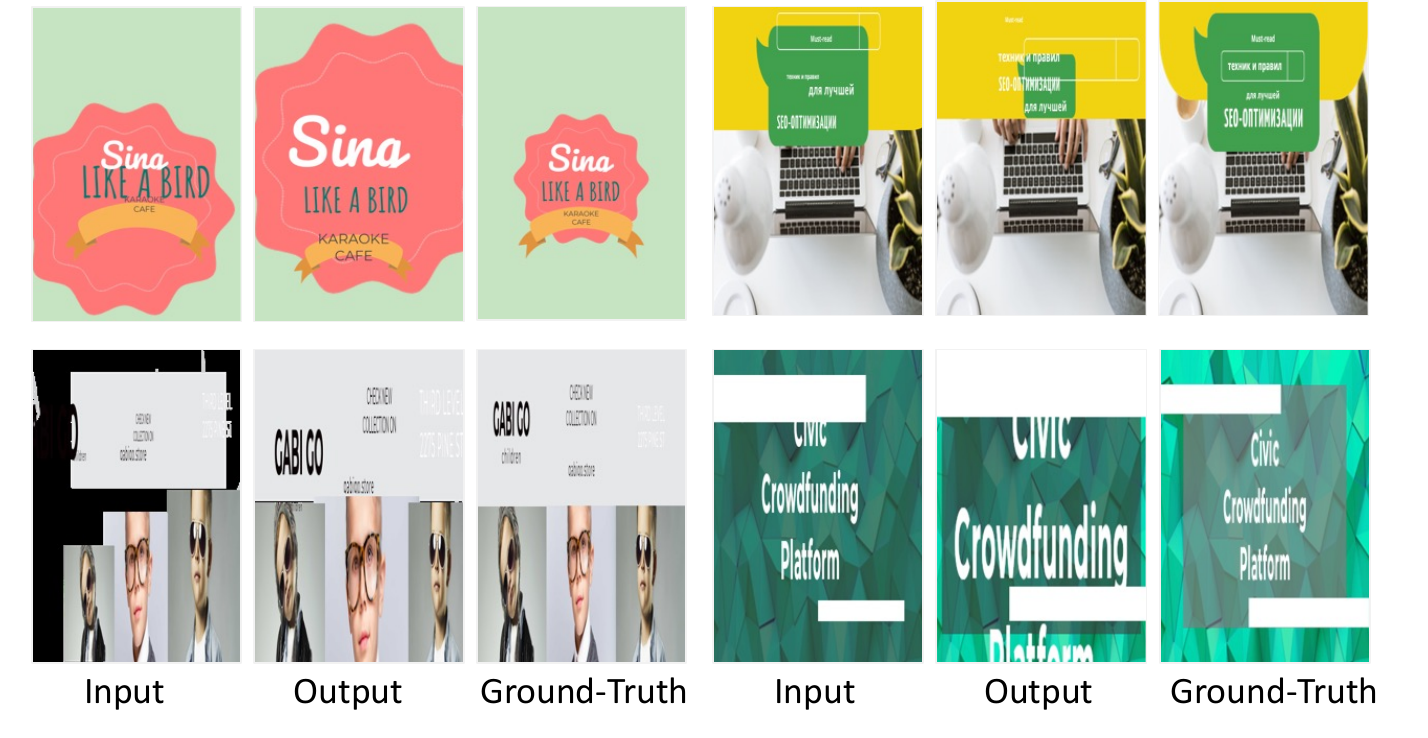}
    \caption{Examples where \ours fails to achieve the optimal refinement.}
    \label{fig:fail}
    \vspace{-10pt}
\end{figure}
We showcase failure cases of \ours in \cref{fig:fail}, ranging from minor to significant failures.
Despite being an excellent scorer and refiner, at-times the signals from the input are weak to correctly guide the layout and scale transformations.

\section{Perturbations used in Dataset Creation}
\begin{figure*}
    \centering
\includegraphics[width=0.99\textwidth]{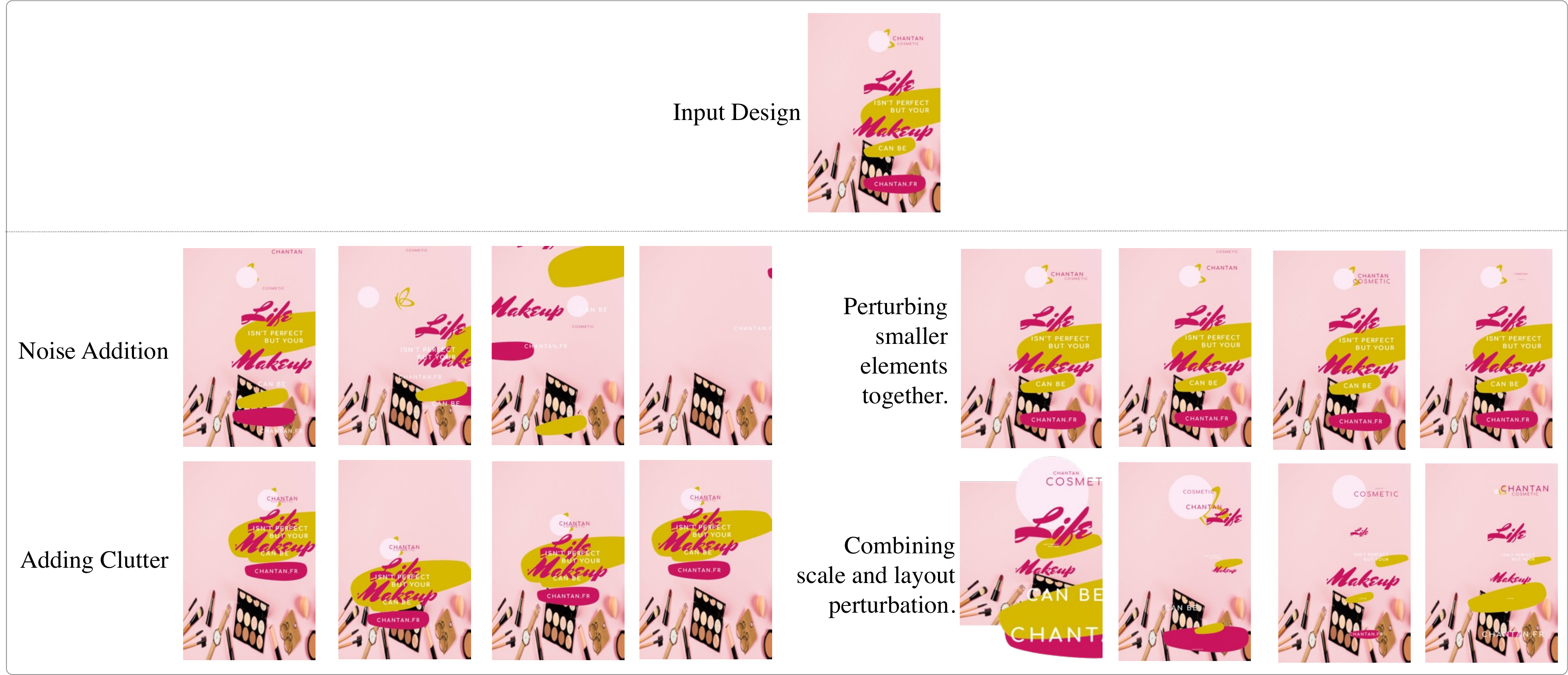}
    \caption{We show the perturbations that we apply to input design, for creating \{good-design, bad-design\} pairs to train our \textit{scorer} model.}
    \label{fig:Perturbations}
\end{figure*}

In \cref{fig:Perturbations}, we show a visualization of the perturbations that we do to the input design to create the dataset to train the \textit{scorer} module, as explained in Sec.~3.1.

\section{Additional Results}
\begin{figure*}
    \centering
\includegraphics[width=0.92\textwidth]{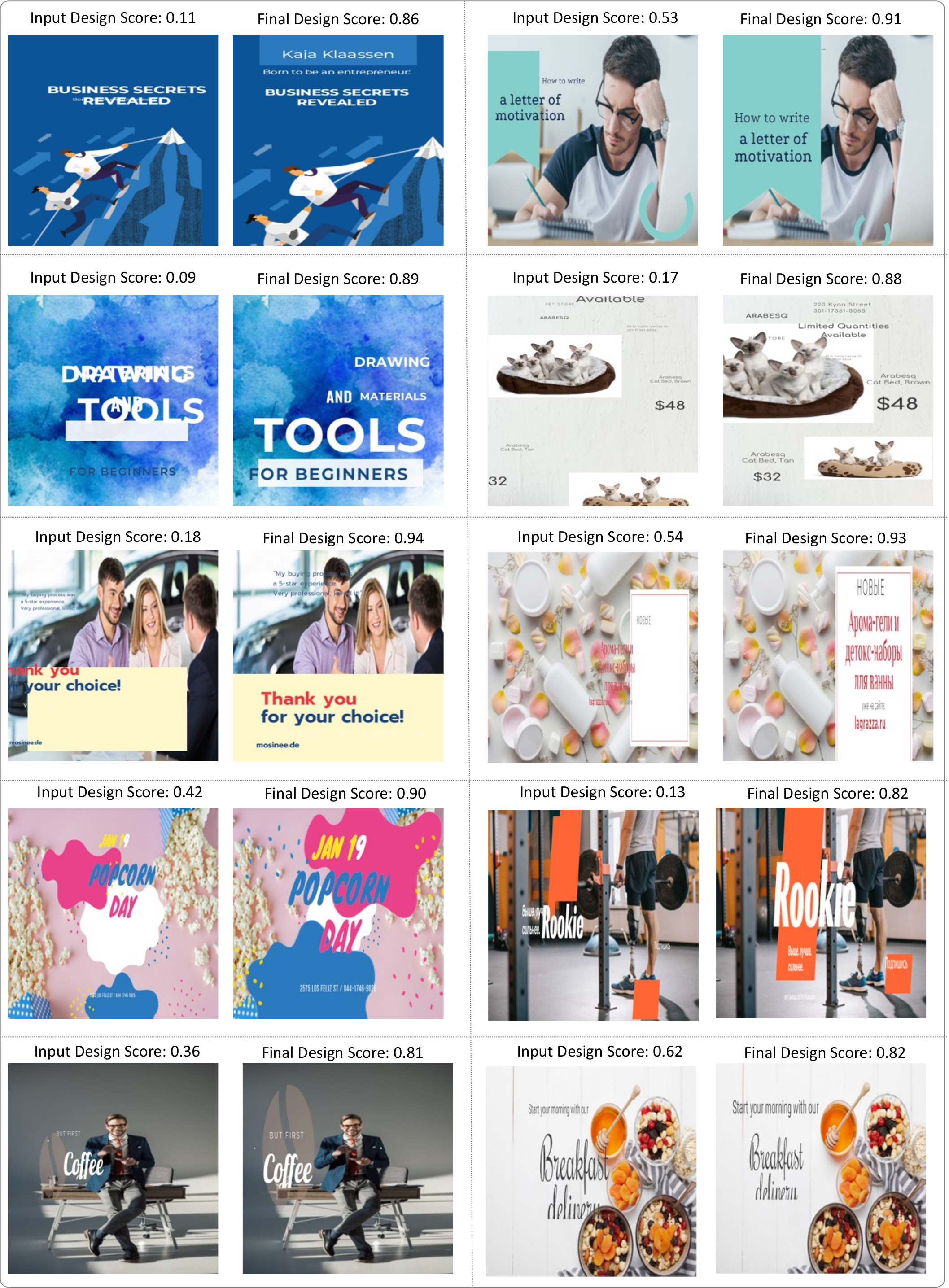}
    \caption{We add more qualitative results here. The input design and its corresponding refined output along with the scores are shown.}
    \label{fig:results1}
\end{figure*}

\begin{figure*}
    \centering
\includegraphics[width=0.92\textwidth]{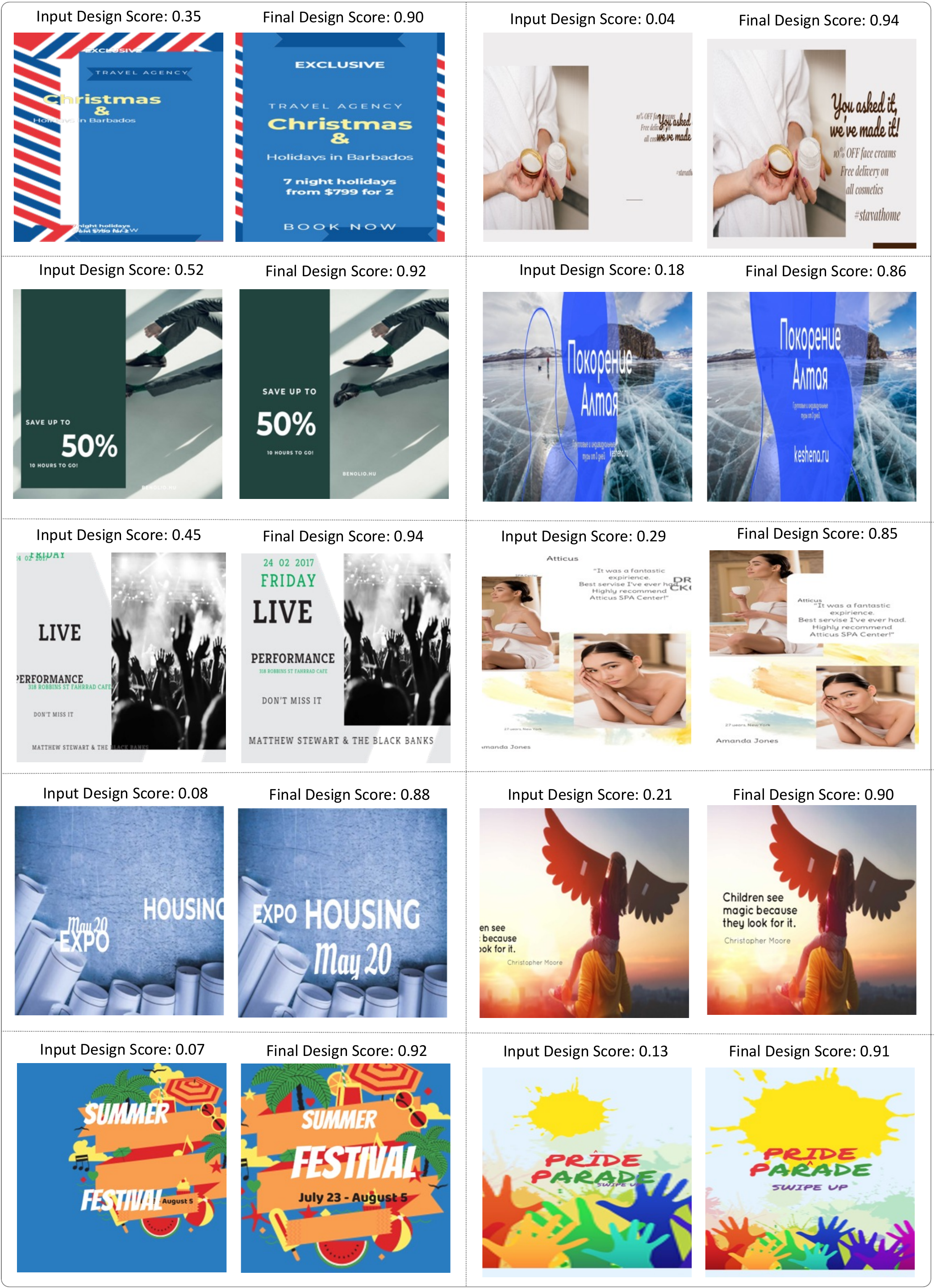}
    \caption{We add more qualitative results here. The input design and its corresponding refined output along with the scores are shown.}
    \label{fig:results2}
\end{figure*}

We include more qualitative results on the scores and refined outputs in \cref{fig:results1,fig:results2}.

\end{document}